\documentclass[runningheads]{llncs}
\usepackage{graphicx}
\usepackage{amssymb}

\usepackage{todonotes}
\usepackage{graphics}
\usepackage{subcaption}
\usepackage{caption}
\usepackage{soul}
\usepackage{csvsimple}
\usepackage{multirow}
\usepackage{float}

\begin{document}
\title{From feature selection to continuous optimization}
\author{Hojjat Rakhshani \and
Lhassane Idoumghar\and
Julien Lepagnot\and
Mathieu Br\'evilliers
}
\authorrunning{H. Rakhshani et al.}
\institute{Universit\'e de Haute-Alsace, IRIMAS-UHA, F-68093 Mulhouse, France}
\maketitle              % typeset the header of the contribution
%

% Automated algorithm design gets more and more important because algorithms get more complex and a developer has to make calls about many aspects, e.g., which subroutine to use or how to set parameters (also known as magic constants). From the perspective of a user, there often exist more than one algorithm to solve a given problem. So, how can we efficiently choose a well performing algorithm? In practice, it is even worse: There exist no single well-performing parameter setting or best algorithm for all kind of possible inputs. Therefore, we have to determine the best settings for new inputs again and again which can be (i) a really time-consuming and tedious task and (ii) a human is often biased by her own experience which is often not optimal.

% papers:

% https://ieeexplore.ieee.org/document/7339682
% https://arxiv.org/pdf/1901.06032.pdf
% Reinforcement learning versus evolutionary computation: A survey onhybrid algorithms
% https://www.quora.com/Is-deep-learning-useful-for-optimization-problems-If-yes-how-does-it-compare-with-traditional-algorithms-such-as-simulated-annealing-evolutionary-algorithms-etc

\begin{abstract}

Metaheuristic algorithms (MAs) have seen unprecedented growth thanks to their successful applications in fields including engineering and health sciences. In this work, we investigate the use of a deep learning (DL) model as an alternative tool to do so. The proposed method, called MaNet, is motivated by the fact that most of the DL models often need to solve massive nasty optimization problems consisting of millions of parameters. Feature selection is the main adopted concepts in MaNet that helps the algorithm to skip irrelevant or partially relevant evolutionary information and uses those which contribute most to the overall performance. The introduced model is applied on several unimodal and multimodal continuous problems. The experiments indicate that MaNet is able to yield competitive results compared to one of the best hand-designed algorithms for the aforementioned problems, in terms of the solution accuracy and scalability.

\end{abstract}
\section{Introduction}

The need for optimization has received a lot of attention in different application areas. Formally, optimization algorithms seek to find a parameter vector $x^{*}$ so as to minimize a cost function $f(x): \mathbb{R}^{D} \rightarrow  \mathbb{R} $, i.e. $f(x^{*}) \leq f(x)$ for all $x \in \Omega $, where $\Omega = {\mathbb{R}}^{D}$ is the search domain and $D$ is the dimension of the problem. There are no a prior hypothesis about $f$ and optimization algorithms should treat them as black-box functions. This motivated the development of MAs which do not take advantages of problem structure. 
\par
MAs are one of the fastest growing fields aimed at solving different complex and highly non-linear real-world problems by inspiration from the process of natural evolution or physical processes~\cite{rakhshani2017snap,kang2018hybrid}. In MAs, we often have a population of candidate solutions that strive for survival and reproduction.  In every iteration,  different search  operators are applied to the candidate solutions and then the population will be updated based on its success in achieving the goal. Over the last decade, there has been an explosion in the development of a variety of extensions to further enhance the performance of MAs. However, there are no clear guidelines on the strengths and weaknesses of alternative methods such as the DL models for developing more enhanced optimization algorithms. 
\par
The DL approaches use a hierarchy  of  features in conjunction with several layers to learn complex non-linear mappings between the input and output layer. As opposite to traditional machine learning methods that use handmade features, the  important features are discovered automatically  and  are  represented  hierarchically. This is known to be  the   strong   point   of  DL  against traditional  machine  learning  approaches. Accordingly, these models have been described as universal learning approaches that are not task specific and can be used to tackle different problems arise in different research domains~\cite{alom2018history}. In this work, we propose a simple, yet effective approach for numerical optimization based on the DL. The proposed MaNet adopts a Convolutional Neural Network (CNN); which are regularized version of fully-connected neural networks inspired from biological visual systems~\cite{krizhevsky2012imagenet}. The "fully-connectedness" of CNNs enables them to tackle the over-fitting problem and it is reasonable to postulate that they may outperform classical neural networks for difficult optimization tasks.
\par
The rest of the paper is organized as follows. Section 2 presents a review on the related works and describes our motivations. In Section 3, we elaborate technical details of the MaNet approach. In Section 4, a series of experiments are conducted to show the performance of the introduced method. The last section summarizes the paper and draws conclusions.

\section{Related works and motivations}

% Figure 1: Figure (a) shows the difference in accuracy with respect to the train size, while Figure (b)

The idea of solving optimization problems using neural networks has an old history which has seen a number of advances in recent years~\cite{kennedy1988neural,snoek2015scalable,amos2017optnet,andrychowicz2016learning,li2016learning}. In~\cite{snoek2015scalable}, authors developed a Bayesian optimization method, called as DNGO, based on deep neural networks for hyperparameter tuning  of large  scale problems with expensive evaluation. The main idea is to  combine large-scale parallelism with an optimization  method to provide an approximate model  of  the  real  cost  function. They show that DNGO scales in a less dramatic fashion compared to the Gaussian process, while maintains its desirable flexibility and characterization of uncertainty. OptNet~\cite{amos2017optnet} is another method proposed for learning optimization tasks by the virtues of DL, sensitivity analysis, bilevel optimization, and implicit differentiation. The authors highlighted the potential power of OptNet networks against existing networks to play mini-Sudoku. In~\cite{andrychowicz2016learning}, researchers investigated automating the design of an optimization algorithm by Long short-term memory deep networks on a number of tasks. Their results outperform hand-designed competitors for simple convex problems, neural network training and  styling images with neural art. Similarly, Li and Malik~\cite{li2016learning} put forward a  deep learning method for automating algorithm design process. They formulate the problem as a reinforcement learning task according to which any candidate algorithm is represented by a policy and the goal is to find an optimal policy. To verify this finding, the authors conducted a set of experiments using different convex and non-convex loss functions correspond to several machine learning models. The obtained results clearly suggest that the automatically designed optimizer converges faster compared to hand-engineered optimizer.
\par
Some of the above mentioned works mainly aim at providing optimal solutions within a very limited computational time~\cite{snoek2015scalable}, while others ~\cite{andrychowicz2016learning,li2016learning} primarily focus on getting better heuristic solutions. These success stories of DL motivated us to investigate the ability of a moderate model so as to make a balance between the solution accuracy and computational time. Altogether, these are the same desired properties in MAs and our work is a step towards investigating the usefulness and strong potential of this research direction.

\section{The proposed method}

This section presents a new optimization method, called MaNet, to explore the possibility of adopting a lightweight deep learning architecture for continuous optimization tasks. In the following, it is assumed that the reader is familiar with the basic concepts of evolutionary computation and deep neural networks. 

\par
The MaNet is designed to have the common properties of the MAs: providing a sufficient good solution with incomplete or imperfect information. It starts  the  optimization  procedure with a set of randomly generated solutions as  genotype. During training the network, MaNet applies the network training components  directly  on  the  genotype, while decodes a genotype into a  phenotype  (i.e., individuals in MAs) only in the last layer. It finds an optimized solution by iteratively improving an initial solution with regard to its cost function. Among different DL models, CNNs trained with an extension of stochastic gradient descent is used to build the MaNet. The CNNs have been central to the largest advances in computer vision~\cite{krizhevsky2012imagenet} and speech processing~\cite{38131}. A CNN is a DL method that uses convolutional layers to filter redundant or even irrelevant input data to increase the performance of the network~\cite{he2016deep}. This consideration also reduces the dimensionality of the input data and  speeds up the learning process in the CNNs. Besides, it allows CNNs to be deeper networks with fewer parameters. Altogether, these properties could make CNNs a potential tool for solving optimization problems; especially when we take into account the history behind the application of feature selection~\cite{rakhshani2019mac} and problem scale reducing~\cite{senjyu2005fast} in the optimization domain.

The architecture of a CNN consists of an input and an output layer, as well as one or more hidden layers. The hidden layers are typically composed of convolutional layers, fully connected layers, normalization layers and pooling layers. The number of hidden layers could be increased depending on the complexities in the input data, but at the cost of more computational expensive simulations.  From the mathematical perspective, convolution layers provide a way of mixing input data with a filter so as to form a transformed feature map. Fully-Connected layers learn non-linear combinations of the high-level features by connecting neurons in one layer to neurons in the previous layer, as seen in multi-layer perceptrons neural networks (MLPs). Moreover, normalization layers are adopted to normalize the data to a network and to speed up learning. This includes batch normalization~\cite{santurkar2018does}, weight normalization~\cite{salimans2016weight}, and layer normalization~\cite{lei2016layer} techniques. Batch normalization is applied to the input data or to the activation of a prior layer, weight normalization is applied to the weights of the layer and layer normalization is applied across the features. The pooling layers are usually inserted in-between successive convolutional layers to further reduce the number of parameters in the network. A CNN network can have local or global pooling layers that may compute a max or an average.

% An example on the functionality of the described layers is given in Figure 1. 

Inspired by the aforementioned components in CNNs, the MaNet is designed to train a model so as to solve an optimization problem (Fig. \ref{fig:MaNet}). The existing feature selection and dimensionality reduction policies in CNNs help MaNet to find complex dependencies between the parameters. The MaNet start optimization by generating a set of random $n \times m$ inputs for the model (i.e., the raw pixel values of the image). So, each individual solution is represented by a matrix rather than a vector. During training the network, convolutional layers transform the initial population layer by layer to a final feasible solution. This large part genotype representation enables the optimizer to keep genetic information that was necessary in the past as a source of exploration, as well as a playground for extracting new features that can be advantageous in the exploitation.

 The MaNet multiplies the initial population with a two-dimensional array of filters that are connected to every disjoint region. The output of multiplying the filters with initial population forms a two-dimensional output array called as "feature map". They are obtained by convolution process upon the initial population with a linear filter, without applying a non-linear function or applying feature normalization methods. Similar to other DL models, the filters/kernels in MaNet are learned using the back-propagation algorithm for each specific optimization task. This is the novel aspect of DL techniques that filter weights are learned during the training of the network and are not hand designed. Accordingly, CNNs are not limited to image data and could be used to extract a variety types of features. Thank to this characteristic, MaNet will be forced to extract the features that are the most important to minimize the loss function for the problem at hand the network is being trained to solve. In each convolution layer, we have some predefined hyperparameters that can be used to modify the behavior of the model: the filter size and the number of filters. The first one simply denotes the dimensions of the filter when applying the convolution process, while the second one determines the number of different convolution filters.
%  ; as given in Fig. \ref{fig:conv-ex}
  \par
 In MaNet, multiple convolution layers are stacked which allows convolution layers to be applied to the output of the previous layer, results in a hierarchically set of more decomposed features. Finally, a Dense layer (or fully-connected) with linear activation function will be used to form the final solution vector. As it can be seen from Fig. \ref{fig:MaNet}, MaNet has a very simple structure and can benefit from the advantage of having a fast network training process\footnote{Netron Visualizer is used to illustrate the model. The tools is available online at: https://github.com/lutzroeder/netron}. Indeed, it has  only 3,742 trainable parameters compared to state-of-the-art models~\cite{simonyan2014very} which have millions or billions of parameters. This could facilitate the application of MaNet for optimization tasks where a small amount of data (i.e., population) is available. 
 
 \begin{figure}[H]
    \centering
    \includegraphics[width=0.80\linewidth]{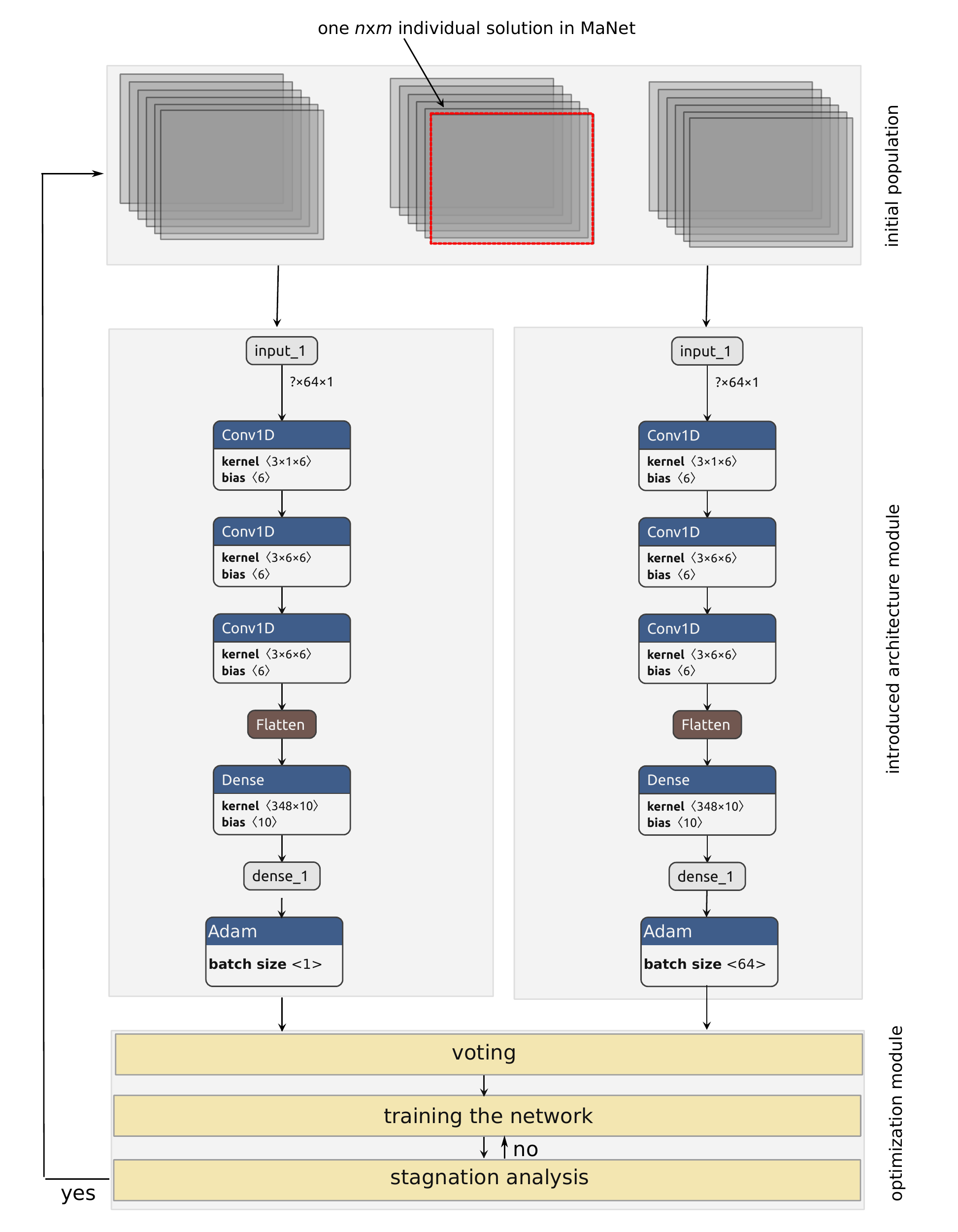}
    \caption{An overview of the proposed optimization architecture. The MaNet is composed of three convolution layers and one Dense layer (or fully connected layer). In each layer, the number of filters and the filter size are 6 and 3, respectively. The activation function for all the layers is proportional to their inputs.}
    \label{fig:MaNet}
\end{figure}

 As it can be seen, the MaNet is composed of two similar architectures which are subjected to different optimization procedures. The first one uses a batch size of one and the other uses 64 as its batch size. The batch size is a hyperparameter of gradient descent that should be tuned for each optimization task. To do so, MaNet integrates a reinforcement strategy inspired from SDCS~\cite{rakhshani2017snap}. Technically speaking, SDCS is a simple metaheuristic algorithm which toggles continually between two snap and drift modes to enhance reinforcement and stability. Based on this idea, MaNet introduces a self-adaptive strategy to tune the batch size hyperparameter.  More precisely, it is looking to see if the best cost function stops improving after some number of epochs, and if so then it restarts the optimization process and continuous the search by the architecture which obtained a higher overall performance so far. Finally, it is worth mentioning to note that the initial population will remain unchanged during training the network and the algorithm will evolve a set of filters. The goal of MaNet then, is to transfer the initial population on one end to evolved solutions on the other hand. This is one of the main differences between MaNet and evolutionary algorithms.

% \begin{figure}[H]
%     \centering
%     \includegraphics[width=0.85\linewidth]{images/Fig4.pdf}
%     \caption{Basic Convolution operation for an $n \times m$ input. The model moves the overlay right one position and repeats the same calculation to get the next result.
%     }
%     \label{fig:conv-ex}
% \end{figure}
% \vspace{-4pt}

% https://hal.inria.fr/hal-01966957/document

\section{Experimental}
\subsection{Experimental setup}
We use a set of 9 benchmark functions given in CEC 2017 [20] to evaluate the performance of the proposed algorithm\footnote{The codes for CEC problems and the jSO algorithm are publicly available at: \\ http://www.ntu.edu.sg/home/EPNSugan/index\_files/CEC2017/CEC2017.htm}. The considered  problems  are  widely  used  in  the optimization community and are challenging for any optimization approach. This work uses several problems  that can  be  classified  into  unimodal  (F1 and F3) and  multimodal (F4-10) minimization functions with   different   properties including separable, non-separable, rotated, ill-condition  and shifted\footnote{F2  has  been  excluded by the organizers because  it  shows  unstable  behavior  especially  for  higher  dimensions~\cite{awad2016problem}}. The aforementioned problems are adopted on the GPU so as to be linked with machine learning libraries. We refer the reader to the  detailed  principle  about  the  definition  of CEC2017 benchmark functions as defined in~\cite{awad2016problem}. To verify the algorithm scalability, 30-dimensional and 50-dimensional problems are used. All functions
should be minimized and have a global minimum at $f(x) = 0$. The results  are  reported  according to their distance  from  the  optimum. We trained MaNet on each problem by using the parallel power of 9 NVIDIA Tesla K20m GPU cards.

It has been shown that various extensions of the differential evolution (DE)~\cite{storn1997differential} algorithm are always among the winners of the CEC competition. Having this is mind, we used jSO~\cite{brest2017single} algorithm for the purpose of comparison which is the second ranked algorithm in CEC2017 competitions for the single objective optimization track. The algorithm is shown to outperform  LSHADE~\cite{tanabe2014improving} (the winner of the CEC2014) and its new extension for CEC2016 (iL-SHADE~\cite{brest2016shade}). All the results are taken from the original study. In  order  to  make  a  fair  comparison,  all the experiment conditions  are  the  same.  The  number  of  function  evaluations  is $10,000 \times D$, where $D$ is the problem dimension~\cite{awad2016problem}. To  tackle  the negative  effects  of  the random  initial  configurations, each  algorithm  were  run  51 times~\cite{awad2016problem}. The  initial  population  is  generated  randomly  within  the  search  bounds  $[-100,100]$. The parameters of the jSO are the same as reported in the original study~\cite{brest2017single}. In MaNet, we have 3 convolution layers which are sequentially connected to each other. In each layer, the number of filters and the filter size are 6 and 3, respectively. The MaNet is a CNN model and needs a lot of input data to be well trained and so the population size is fixed to $n=5,000$. Moreover, $m$ is considered to be 64 for all the problems. The MaNet will be optimized using the Adam algorithm~\cite{kingma2014adam}. 
\vspace{-0.5cm}
\subsection{Results and discussion}

Tables   1-2   present   best,   worst,   mean   and   standard   deviation  (Std.)  results  of the MaNet and jSO on 9 problems over 51 runs. Table 1 reports the results for 30 dimensional problems, while Table 2 shows the performance of the competitive algorithms for 50 dimensional cases. In  these  tables,  a statistical test is  also  presented  to  assess  the  significance  of  performance  between the results  of  the  jSO  and  MaNet.

\vspace{-0.35cm}

\begin{table}[H]
\centering 

\caption{The obtained results by MaNet and jSO for 30 dimensional problems over 51 runs~\cite{awad2016problem}. The results for jSO are directly taken from the original paper~\cite{brest2017single}.}
\scriptsize
\begin{tabular}{|l|l|l|l|l|l|l|c|}
\hline
Function            & Algorithm & Best       & Worst      & Mean                & Median     & Std.       & Sign                  \\ \hline
\multirow{2}{*}{1}  & MaNet     & $3.71e+02$ & $1.33e+03$ & $7.94e+02$          & $8.02e+02$ & $2.03e+02$ & \multirow{2}{*}{$-$} \\ \cline{2-7}
                    & jSO       & $0.00e+00$ & $0.00e+00$ & \textbf{0.00e+00} & $0.00e+00$ & $0.00e+00$ &                    \\ \hline
\multirow{2}{*}{3}  & MaNet     & $3.69e+04$ & $7.10e+04$ & $5.85e+04$          & $5.85e+04$ & $6.46e+03$ & \multirow{2}{*}{$-$} \\ \cline{2-7}
                    & jSO       & $0.00e+00$ & $0.00e+00$ & \textbf{0.00e+00} & $0.00e+00$ & $0.00e+00$ &                    \\ \hline
\multirow{2}{*}{4}  & MaNet     & $1.46e-05$ & $3.99e+00$ & \textbf{5.88e-01} & $6.79e-04$ & $1.41e+00$ & \multirow{2}{*}{$+$} \\ \cline{2-7}
                    & jSO       & $5.86e+01$ & $6.41e+01$ & $5.87e+01$          & $5.86e+01$ & $7.78e-01$ &                    \\ \hline
\multirow{2}{*}{5}  & MaNet     & $0.00e+00$ & $1.99e+00$ & \textbf{5.85e-01} & $1.34e-07$ & $6.59e-01$ & \multirow{2}{*}{$+$} \\ \cline{2-7}
                    & jSO       & $3.98e+00$ & $1.32e+01$ & $8.56e+00$          & $8.02e+00$ & $2.10e+00$ &                    \\ \hline
\multirow{2}{*}{6}  & MaNet     & $0.00e+00$ & $0.00e+00$ & \textbf{0.00e+00} & $0.00e+00$ & $0.00e+00$ & \multirow{2}{*}{$=$} \\ \cline{2-7}
                    & jSO       & $0.00e+00$ & $0.00e+00$ & \textbf{0.00e+00} & $0.00e+00$ & $0.00e+00$ &                    \\ \hline
\multirow{2}{*}{7}  & MaNet     & $3.26e+01$ & $3.41e+01$ & \textbf{3.33e+01} & $3.33e+01$ & $3.91e-01$ & \multirow{2}{*}{$+$} \\ \cline{2-7}
                    & jSO       & $3.61e+01$ & $4.31e+01$ & $3.89e+01$          & $3.91e+01$ & $1.46e+00$ &                    \\ \hline
\multirow{2}{*}{8}  & MaNet     & $0.00e+00$ & $4.97e+00$ & \textbf{2.29e+00} & $1.99e+00$ & $1.15e+00$ & \multirow{2}{*}{$+$} \\ \cline{2-7}
                    & jSO       & $4.97e+00$ & $1.30e+01$ & $9.09e+00$          & $8.96e+00$ & $1.84e+00$ &                    \\ \hline
\multirow{2}{*}{9}  & MaNet     & $0.00e+00$ & $0.00e+00$ & \textbf{0.00e+00} & $0.00e+00$ & $0.00e+00$ & \multirow{2}{*}{$=$} \\ \cline{2-7}
                    & jSO       & $0.00e+00$ & $0.00e+00$ & \textbf{0.00e+00} & $0.00e+00$ & $0.00e+00$ &                    \\ \hline
\multirow{2}{*}{10} & MaNet     & $1.09e+04$ & $1.13e+04$ & $1.11e+04$          & $1.11e+04$ & $1.19e+02$ & \multirow{2}{*}{$-$} \\ \cline{2-7}
                    & jSO       & $1.04e+03$ & $2.04e+03$ & \textbf{1.53e+03} & $1.49e+03$ & $2.77e+02$ &                    \\ \hline
\end{tabular}
\end{table}

\vspace{-0.8cm}

% \usepackage{multirow}
\begin{table}[H]
\centering

\caption{The obtained results by MaNet and jSO for 50 dimensional problems over 51 runs~\cite{awad2016problem}. The results for jSO are directly taken from the original paper~\cite{brest2017single}.}
\scriptsize
\begin{tabular}{|l|l|l|l|l|l|l|c|}
\hline
Function            & Algorithm & Best       & Worst      & Mean       & Median     & Std.       & Sign               \\ \hline
\multirow{2}{*}{1}  & MaNet     & $3.67e+02$ & $2.06e+03$ & $1.39e+03$ & $1.46e+03$ & $3.71e+02$ & \multirow{2}{*}{$-$} \\ \cline{2-7}
                    & jSO       & $0.00e+00$ & $0.00e+00$ & \textbf{0.00e+00} & $0.00e+00$ & $0.00e+00$ &                    \\ \hline
\multirow{2}{*}{3}  & MaNet     & $9.80e+04$ & $1.42e+05$ & $1.23e+05$ & $1.25e+05$ & $8.88e+03$ & \multirow{2}{*}{$-$} \\ \cline{2-7}
                    & jSO       & $0.00e+00$ & $0.00e+00$ & \textbf{0.00e+00} & $0.00e+00$ & $0.00e+00$ &                    \\ \hline
\multirow{2}{*}{4}  & MaNet     & $3.10e-06$ & $1.53e-03$ & \textbf{8.22e-04} & $9.96e-04$ & $4.46e-04$ & \multirow{2}{*}{$+$} \\ \cline{2-7}
                    & jSO       & $1.32e-04$ & $1.42e+02$ & $5.62e+01$ & $2.85e+01$ & $4.88e+01$ &                    \\ \hline
\multirow{2}{*}{5}  & MaNet     & $1.99e+00$ & $1.09e+01$ & \textbf{6.15e+00} & $5.97e+00$ & $2.20e+00$ & \multirow{2}{*}{$+$} \\ \cline{2-7}
                    & jSO       & $8.96e+00$ & $2.39e+01$ & $1.64e+01$ & $1.62e+01$ & $3.46e+00$ &                    \\ \hline
\multirow{2}{*}{6}  & MaNet     & $0.00e+00$ & $0.00e+00$ & \textbf{0.00e+00} & $0.00e+00$ & $0.00e+00$ & \multirow{2}{*}{$=$} \\ \cline{2-7}
                    & jSO       & $0.00e+00$ & $0.00e+00$ & \textbf{0.00e+00} & $0.00e+00$ & $0.00e+00$ &                    \\ \hline
\multirow{2}{*}{7}  & MaNet     & $5.49e+01$ & $5.65e+01$ & \textbf{5.58e+01} & $5.59e+01$ & $3.62e-01$ & \multirow{2}{*}{$+$} \\ \cline{2-7}
                    & jSO       & $5.75e+01$ & $7.42e+01$ & $6.65e+01$ & $6.66e+01$ & $3.47e+00$ &                    \\ \hline
\multirow{2}{*}{8}  & MaNet     & $1.99e+00$ & $8.95e+00$ & \textbf{5.41e+00} & $5.97e+00$ & $1.99e+00$ & \multirow{2}{*}{$+$} \\ \cline{2-7}
                    & jSO       & $9.95e+00$ & $2.41e+01$ & $1.70e+01$ & $1.70e+01$ & $3.14e+00$ &                    \\ \hline
\multirow{2}{*}{9}  & MaNet     & $0.00e+00$ & $0.00e+00$ & \textbf{0.00e+00} & $0.00e+00$ & $0.00e+00$ & \multirow{2}{*}{$=$} \\ \cline{2-7}
                    & jSO       & $0.00e+00$ & $0.00e+00$ & \textbf{0.00e+00} & $0.00e+00$ & $0.00e+00$ &                    \\ \hline
\multirow{2}{*}{10} & MaNet     & $1.86e+04$ & $1.88e+04$ & $1.87e+04$ & $1.87e+04$ & $6.25e+01$ & \multirow{2}{*}{$-$} \\ \cline{2-7}
                    & jSO       & $2.40e+03$ & $3.79e+03$ & \textbf{3.14e+03} & $3.23e+03$ & $3.67e+02$ &                    \\ \hline
\end{tabular}
\end{table}

The  results  of  the  Wilcoxon  rank  sum  test   are  reported  at  the  $95\%$  confidence  level.  In  these   tables,   ‘+’   shows   that   MaNet   significantly  outperforms  the  jSO with  $95\%$  certainty;   ‘-’   indicates   that   the  jSO   is   significantly  better  than  MaNet;  and  ‘=’ shows there is no statistical different between the two compared algorithms. The  significant  results  are  given  in  bold.  For  further  validation,   convergence   graphs of jSO and MaNet for 30 dimensional functions F4 and F8 are given in Fig. \ref{fig:conv-graph}.

As can be seen from Tables 1-2, jSO gives more accurate solutions for the unimodal benchmarks F1 and F3 for both 30-dimensional and 50-dimensional cases. Moreover, with the exceptions of F10, MaNet has equal or significantly better performance on all the multimodal benchmark functions. In fact, the results indicate that MaNet significantly outperforms the jSO on 4 functions (F4-F8), obtains an equal performance on 2 functions (F6 and F9), and has worst results on 3 test cases (F1, F3 and F10). Furthermore, we can see that MaNet is a robust algorithm according to the reported standard deviation results. In addition, these experimental results have confirmed that MaNet is not very sensitive to the increment of dimension and is scalable.  Considering  Fig. \ref{fig:conv-graph}, it can be seen also that  MaNet has  a  more  rapid  convergence  rate than  the  jSO  algorithm for function F4 and F8. In MaNet, we assume that not selection, but rather the combination of different filters is the main source of evolution and that is the reason for having unstable convergence behavior on these functions.  

Altogether, these promising results have confirmed that MaNet has a competitive results in comparison with one of the best designed algorithm for the CEC2017 problems. This is quite interesting because MaNet doesn't borrow any search strategy or components from the previously proposed methods for the CEC problems; including CMAES~\cite{loshchilov2013cma}, DE, jADE~\cite{zhang2009jade}, SADE~\cite{qin2005self}, SHADE~\cite{tanabe2013success}, L-SHADE~\cite{tanabe2014improving}, i-LSHADE~\cite{brest2016shade} and jSO.

\begin{figure}
\begin{subfigure}{.53\textwidth}
  \centering
  \includegraphics[width=\linewidth]{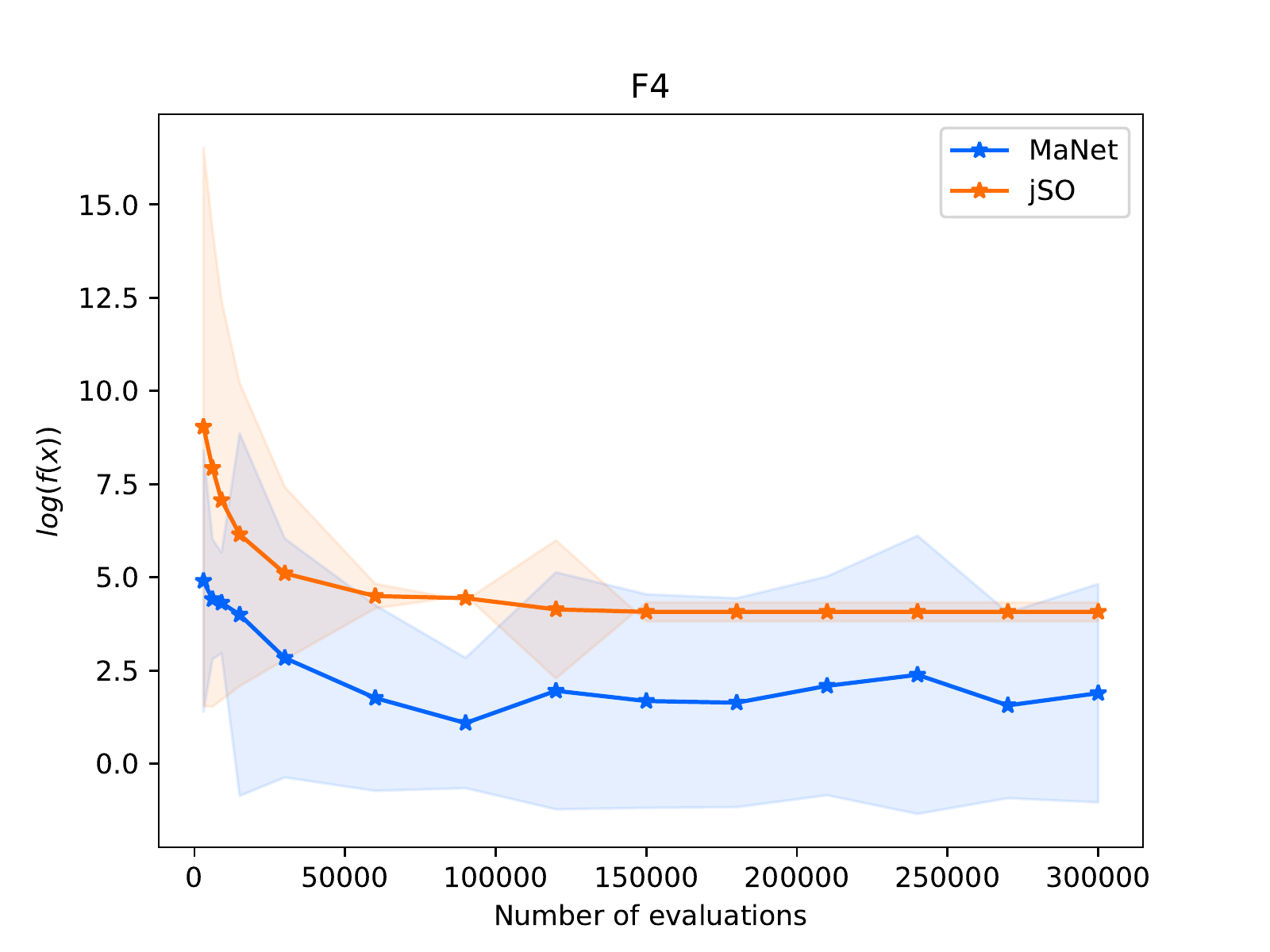}

  \label{fig:sfig1}
\end{subfigure}%
\begin{subfigure}{.53\textwidth}
  \centering
  \includegraphics[width=\linewidth]{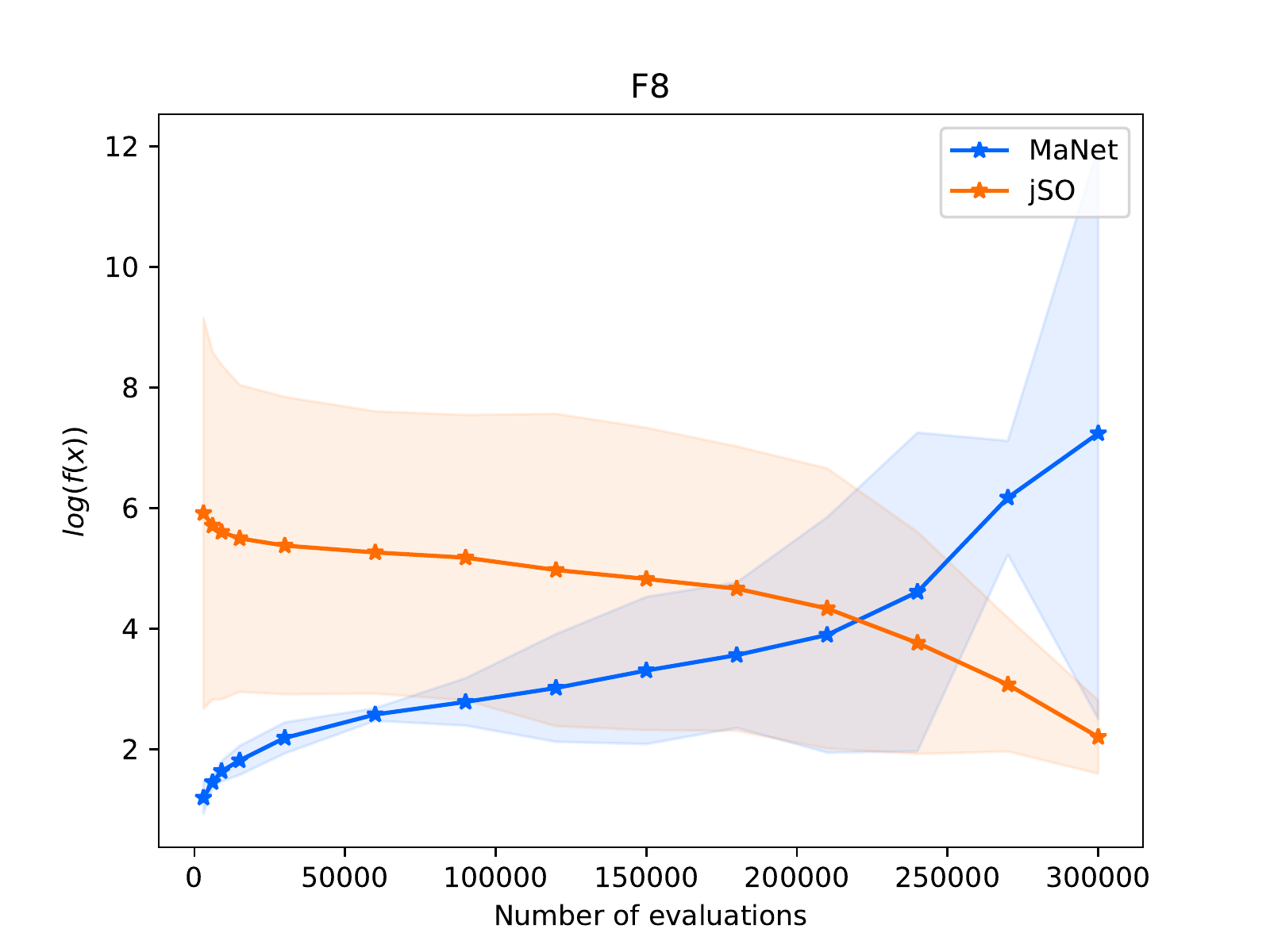}

  \label{fig:sfig2}
\end{subfigure}
\setcounter{figure}{2}  
\centering
\caption{Convergence graphs of the jSO and MaNet for 30 dimensional functions F4 and F8 over 51 runs}
\label{fig:conv-graph}
\end{figure}

As a future work, we are intended to apply the proposed MaNet to all the problems over all the dimensions. Besides, we have to find a way in order to adjust the learning rate hyperparameter for each problem. From Fig. \ref{fig:conv-graph} one can see that a high learning rate in Adam causes the network to generate large numbers for F8 and the updates are going to be just as large. After that, we would like to apply the proposed methodology to more complicated real-world optimization problems.

% \begin{figure*}
% \begin{subfigure}{.50\textwidth}
%   \centering
%   \includegraphics[width=\linewidth]{images/Fig2_10.pdf}
%   \caption{(a)}
%   \label{fig:sfig1}
% \end{subfigure}%
% \begin{subfigure}{.50\textwidth}
%   \centering
%   \includegraphics[width=\linewidth]{images/Fig2_50.pdf}
%   \caption{(b)}
%   \label{fig:sfig2}
% \end{subfigure}
% \setcounter{figure}{1}  
% \caption{Figure (a) shows the diverging dot plot between L-SHADE and MaNet algorithms for 10 dimensions, while Figure (b) presents the results for 50 dimensions}
% \label{fig:acc-vs-train-size}
% \end{figure*}

\section{Conclusion}
This study proposed a new optimization algorithm based on the DL in order to provide an improved search
process. The proposed method verifies convergence conditions by using a CNN model. The simple structure of the MaNet along with feature selection and dimension reduction strategies result in an architecture at a relatively low computational cost. The MaNet optimizer is evaluated using unimodal and multimodal optimization benchmarks from CEC2017 test suite. The obtained results are statistically analyzed and compared with state-of-the-art jSO algorithm.  Evaluations confirm that the introduced MaNet optimization model has a competitive performance in terms of the final solution accuracy and scalability compared to one of the best designed algorithms for the problem at hand. 

\section*{Acknowledgments}
This research was supported through computational resources provided by M\'esocentre of Strasbourg: https://services-numeriques.unistra.fr/
\bibliographystyle{splncs04}
\bibliography{samplepaper}

\end{document}